# Informing Artificial Intelligence Generative Techniques using Cognitive Theories of Human Creativity

Steve DiPaola[1], Liane Gabora[2], and Graeme McCaig[1]

[1] Simon Fraser University, Vancouver BC, Canada
sdipaola@sfu.ca, gmccaig@sfu.ca
[2] University of British Columbia, Kelowna BC, Canada
liane.gabora@ubc.ca

**Abstract:** The common view that our creativity is what makes us uniquely human suggests that incorporating research on human creativity into generative deep learning techniques might be a fruitful avenue for making their outputs more compelling and human-like. Using an original synthesis of Deep Dream-based convolutional neural networks and cognitive based computational art rendering systems, we show how honing theory, intrinsic motivation, and the notion of a "seed incident" can be implemented computationally, and demonstrate their impact on the resulting generative art. Conversely, we discuss how explorations in deep learning convolutional neural net generative systems can inform our understanding of human creativity. We conclude with ideas for further cross-fertilization between AI based computational creativity and psychology of creativity.

## 1 Introduction

It is often said that what makes us distinctively human is our capacity for great works of art and science: in short, our creativity. Yet although some computational creativity research has taken its cue from the psychology of creativity (e.g., Wiggins 2006; Gabora & DiPaola, 2012), computational creativity largely proceeds independently of psychological research on creativity. This paper attempts to bridge this divide by exploring how key concepts from the psychology of creativity can revitalize computational creativity research, and vice versa, focusing on the domain of generative art.

First, we summarize work on Deep Dream (DD) (Mordvintsev et al. 2015), an algorithm based on deep-learning convolutional neural networks (CNNs) (Krizhevsky et al. 2012), that blends visual qualities from multiple source images to create a new output image. DD and its deep learning-based variants have gained notoriety and engendered public speculation regarding the extent to which it is truly creative, and its potential to replace human artists. Next, we show how some concepts from the psychology of creativity have been explored in the DD environment. Finally, we suggest approaches from the psychology of creativity that might inspire further generative art developments.

## 2  Neural Net Image Generation with DeepDream: Computational Perspective

### 2.1  Deep Learning Convolutional Neural Nets

Neural networks lend themselves to modeling creative abilities such as concept combination because of their capacity to incorporate context and forge associations (Boden 2004). Deep learning is a collection of network-based machine learning methods that are notable for their performance in tasks such as object recognition (Krizhevsky et al. 2012), as well as their similarities to aspects of human vision and brain function (DiCarlo et al. 2012). Running input data through a trained network models perception, and some network types are capable of using feedback connections to create novel data generalized from what has been learned (Salakhutdinov and Hinton 2009; Ranzato and Hinton 2010), in a manner that has been likened to hallucination (Reichert et al. 2013).

The convolutional neural network (CNN) (LeCun et al. 1998) is a deep feedforward neural net architecture usually trained with backpropagation. It obtains good generalization, efficient training, and good invariance to input distortions, by incorporating reasonable assumptions about the input image domain through mechanisms of local receptive fields, shared weights, and sub-sampling.

### 2.2  Principles of the DeepDream Algorithm

We now review the principles of DeepDream (Mordvintsev et al. 2015) as they pertain to the present exploration of the creative process, focusing on DD's guided mode. DD uses a trained CNN to generate a transformed version of a source image, emphasizing certain visual (semantic and/or stylistic) qualities, as illustrated in Figure 1. The algorithm begins by analyzing a human-selected guide image, which is propagated from the lowest (pixel) layer to a chosen higher layer. Higher layers encode the image in terms of progressively more abstract features. This encoding is stored as a guide-feature vector. Next, the algorithm initializes the set of pixels that constitute an updated source image which we refer to as the canvas image. The canvas image is gradually transformed into the output image. The canvas image is propagated up through the network to the same layer as the guide-feature vector, yielding a canvas-feature vector.

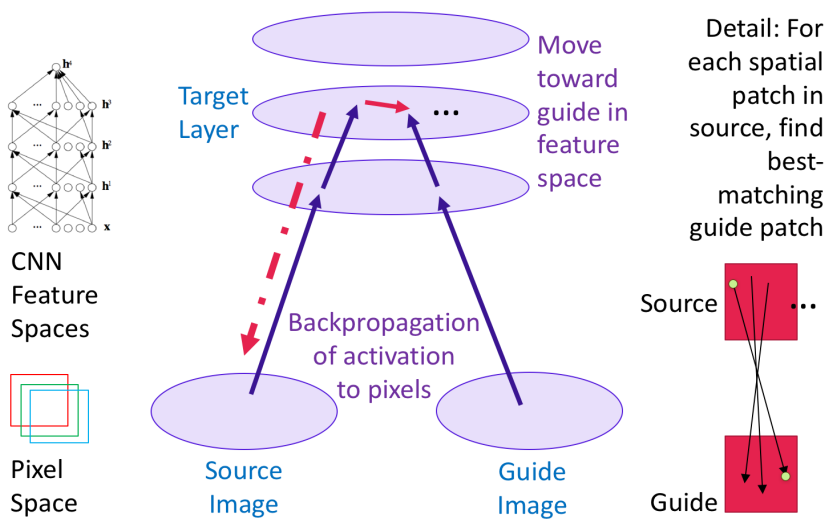

**Fig 1.** The Deep Dream algorithm.

A loss function is defined to measure the difference between the two feature vectors. (The feature vectors are separated into spatial patches, and features from each spatial patch in the source are compared to the best-matching spatial patch in the guide. Thus, a dreamed output image tends to pick up on certain types of shape and texture, found within the guide image, that bear similarity to its own shapes and textures.) From this loss function, a gradient is found and back-propagated back to the lowest (pixel) layer. A small step in the direction thus found is applied as a change to pixel values. Repeating the cycle (of update pixels. propagate forward, find gradient, propagate downward) constitutes a gradient ascent process.

Note that the original guided DD algorithm maximizes the dot product between source and guide features. This can result in textures and patterns that go beyond the guide features—the dotted line in Figure 1—into exaggerated or "hallucinatory" appearances. In contrast, for some of the work here, we have introduced variations on the algorithm, which instead minimize the distance between source and guide features. This tends to produce "tamer" output imagery with textures that preserve the look of the guide image.

## 2.3 Blends and 2-phase Aspects of Deep Dream

To create a blend of two images, DD first generalizes each image by propagating it through a deep CNN and representing it according to the resulting tensor encoding at a certain network layer(s). Depending on the height and type of the network layer, the given encoding analyzes the image according to a particular set of visual features. Much as in the parable of the blind men who each describe an elephant in a different way, different layers "see the image" different ways. Thus, the visual blend depends on points of similarity between guide and source, as viewed from the perspective of a certain network layer encoding. In turn, the nature of a layer encoding depends on both the network architecture and the original training data which caused features to develop as a form of long-term memory.

To enhance similarity between the source image and the guide, the algorithms use a reiterative two-phase creative process of alternating divergence and convergence: any similarities found at a high/abstract level are manifested back at the pixel level.

## 2.4 Modifying DeepDream towards Art Understanding and Generation

When investigating cognitive based human creativity theory, one of the most significant set of artifacts of human creativity is the history of fine art paintings. We have amassed, from many sources, a large 52 GB database of high resolution and labeled fine art paintings and drawings from the 15th century through to present day. While most DD systems used networks pre-trained on object recognition data such as ImageNet (Russakovsky et al. 2015), we use our own modified DeepDream code, and train new models using different subsets of 15th century-onward paintings and drawings. We identified a major hurdle for the use of regular CNN training techniques for this task. Most artists make under 200 large paintings in their lifetime, Picasso being an outlier with almost 2000. There is insufficient visual material for CNN training.

We overcame this problem using what we call a "tight style and tile" method. In many visual deep learning applications, the goal is something like detecting and identifying objects according to category (e.g. human, animal, car, or perhaps more finely such as species or model). For visual art/creativity, this goal is less important; what does matter is artistic style (visual style, stroke style, texture style, color palette style). Therefore, we developed a hierarchical tiling cut method to cut up each artwork into 50+ tiles of a variety of sizes. This gives us much more material per artist to work with, and much better results. We created human-labelled art style categories that cluster quite tightly, e.g., a category might include only early Van Gogh's with one similar style. The 50 or so images from such a category can be stochastically cut into approximately 2000 tiles. This gives strong deep learning metrics, and superior results. This Stochastic Tiling Technique makes it possible to train on the style of a fine art painting as opposed to object detection which, as we discuss below, provides an inroad for the exploration of personal creative style.

## 3 Analyzing DeepDream via Creativity and Psychological Theory

This section provides examples of images processed by the DD algorithm, and discusses them in terms of creative cognitive mechanisms. Certain characteristics of DeepDream output are particularly evident in Figure 2. Firstly, the algorithm does not merely superimpose layer-specific features at random over the image; features tend to be emphasized and grown starting from those image regions that already contain said features. For example, a layer that emphasizes circle and arc shapes tends to place them in pre-existing arc-shaped parts of the image, such as the curved orbital region around a subject's eye. On the other hand, if the algorithm is run for many iterations, all image regions are eventually forced in the direction of a high-activating feature, essentially making something out of nothing.

Another striking aspect of DD output is that the textures and shapes convey a sense of completion and continuation, or flow. Figure 2 shows two images which, despite being quite different, have certain similarities, including a recognizable subject and a repeated pattern composed of elements of similar size and shape, not overlapping or interrupted. This arises from the type of optimization performed by the search process; the total activation of a layer is more enhanced when neighboring features work together without overlapping or disrupting each other.

It is possible with DD to achieve a specific desired kind of style, through thoughtful selection and trial of promising guide images; see for example the organic, nature-based style achieved using nature-image guides illustrated in Figure 3. To demonstrate the range of styles possible, Figure 4a shows a sketch-style 2D image of a 3D avatar.

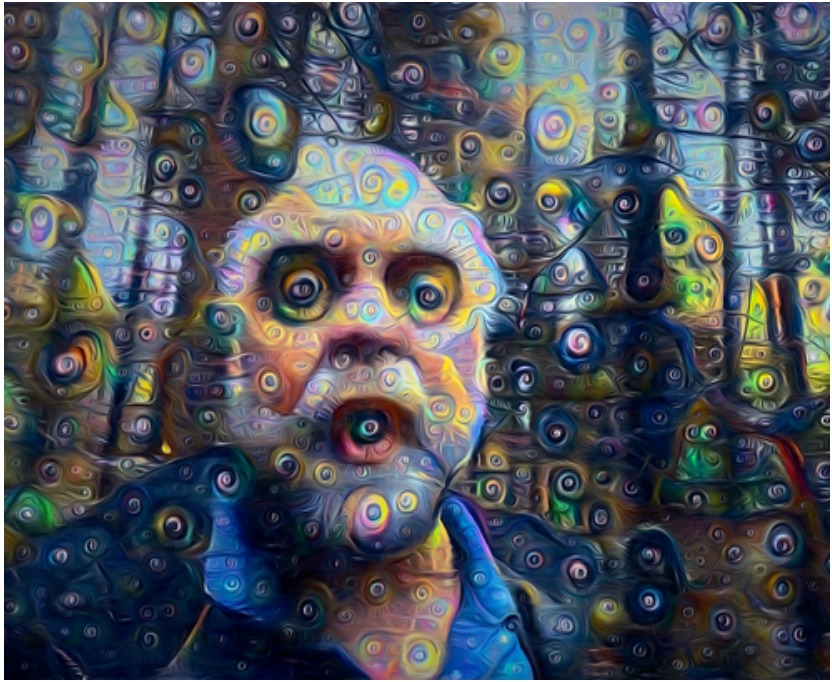
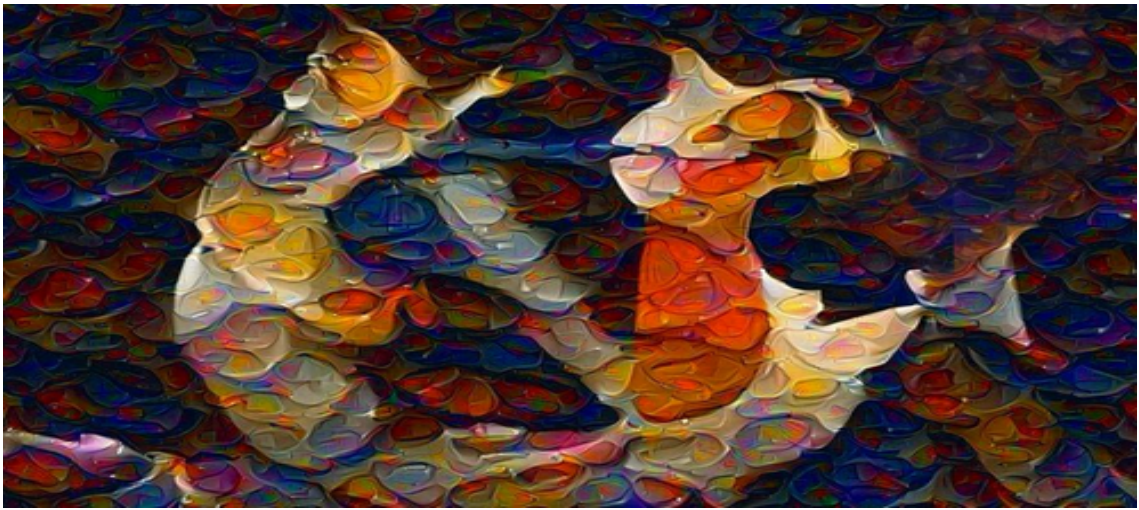

**Fig 2.** Two quite different illustrations of some of DD's most typical features, including an exaggerated style composed of repeating structures.

In our exploration of the DD algorithm, we have situated DD within the wider scope of an artificial painter system. DD (or similar systems) can play the role of the artists' perception and imagination (for abstraction, narrative, and emphasis), while a further artistic painting algorithm models the step that occurs in manifesting imagery onto the canvas and dialoging with the artist's materials. It would make sense to have a cyclical interaction between the perception/imagination phase and the stroke-painting phase, which is a research direction we are presently investigating. In our first attempts we have applied the DD module along with a separate stroke-based painterly rendering module in alternating passes, which already gives quite pleasing results (as in Fig 4b).

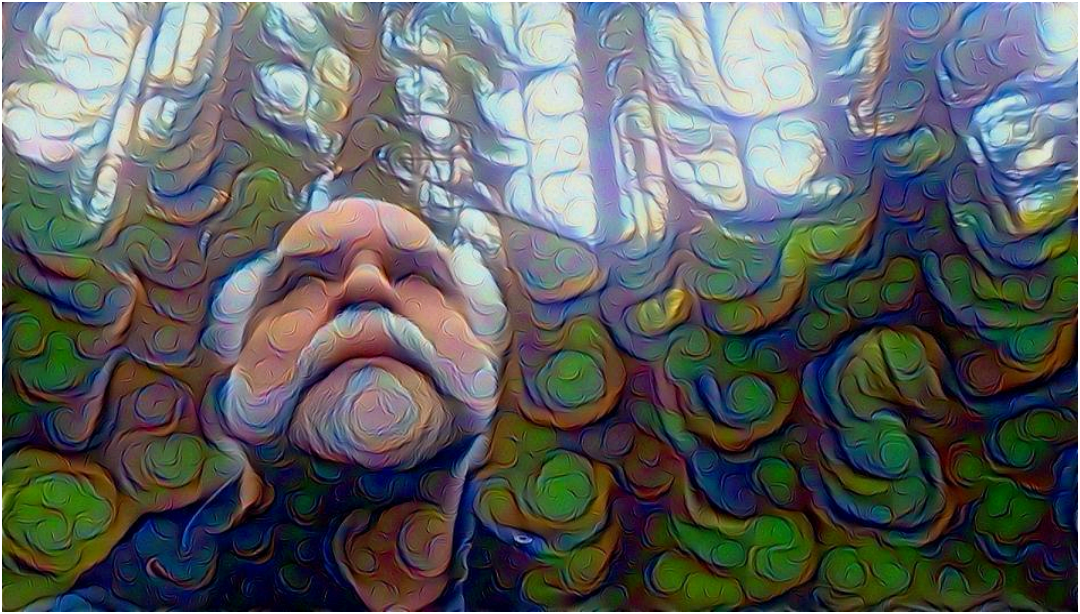

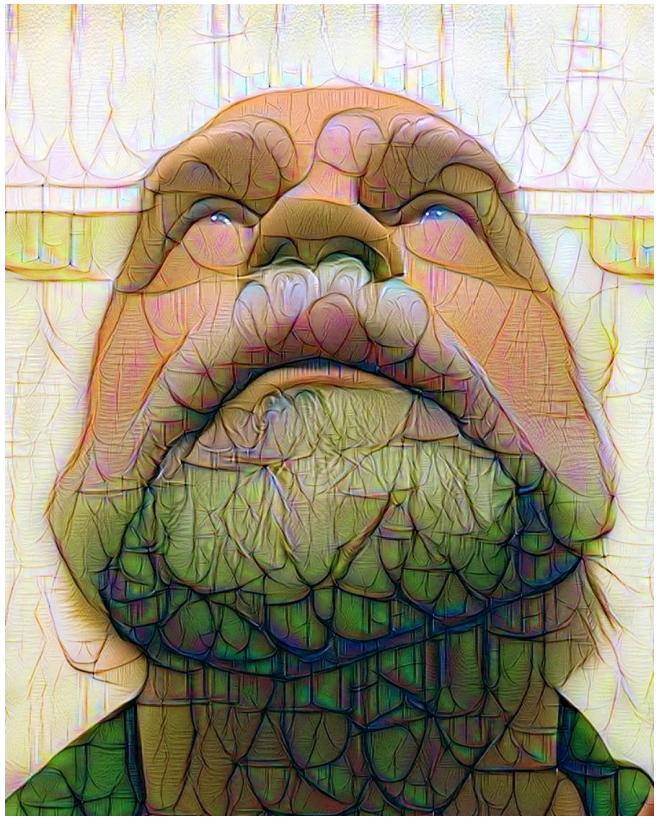

**Fig 3.** Two pieces that illustrate the more organic style achieved using nature guides.

## 3.1 Generating and Evaluating Novelty and Aesthetic Value

We can also analyze DD in terms of other ideas about computational creativity — to what extent can DD and related algorithms pursue generated outputs having novelty and value (widely accepted as the defining characteristics of creative production). We first inquire whether DD makes explicit autonomous evaluations of the novelty and value of its output. This is important for creativity: for example, Jennings (2010) includes such autonomous evaluation as a necessary condition for creative autonomy. (He also includes ability to change evaluative standards, a condition not met by DD on its own). In terms of novelty, when used as standalone systems, DD does not maintain or optimize explicit measures of novelty as an image is produced. In terms of value, the situation is more favorable: DD isolates and maximizes subsets of the network features evoked by each input image, resulting in the maintenance or enhancement of certain aspect image qualities (or whole levels of abstraction) at the expense of other qualities. This process can be construed as the computation of one or more aesthetic value metrics, and bears a resemblance to neuroaesthetic principles of art, such as Zeki's (2001) notions of highlighting/stimulating discrete portions of visual processing and translating the brain's abstractions on to the canvas or Ramachandran and Hirstein's (1999) laws of peak-shift and isolation. In addition, Ramachandran and Hirstein's claim that visual metaphor adds to aesthetic appeal is a clear connection between DD and aesthetic value, and the tendency of DD to create well-formed, complete-looking visual shapes may also contribute to aesthetic value.

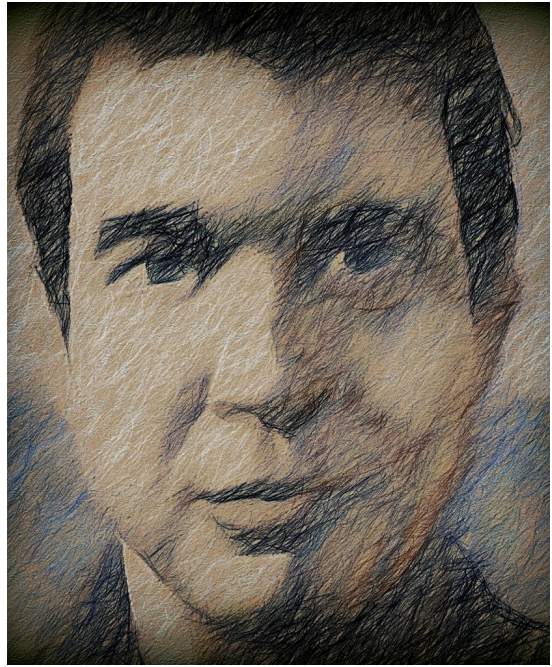

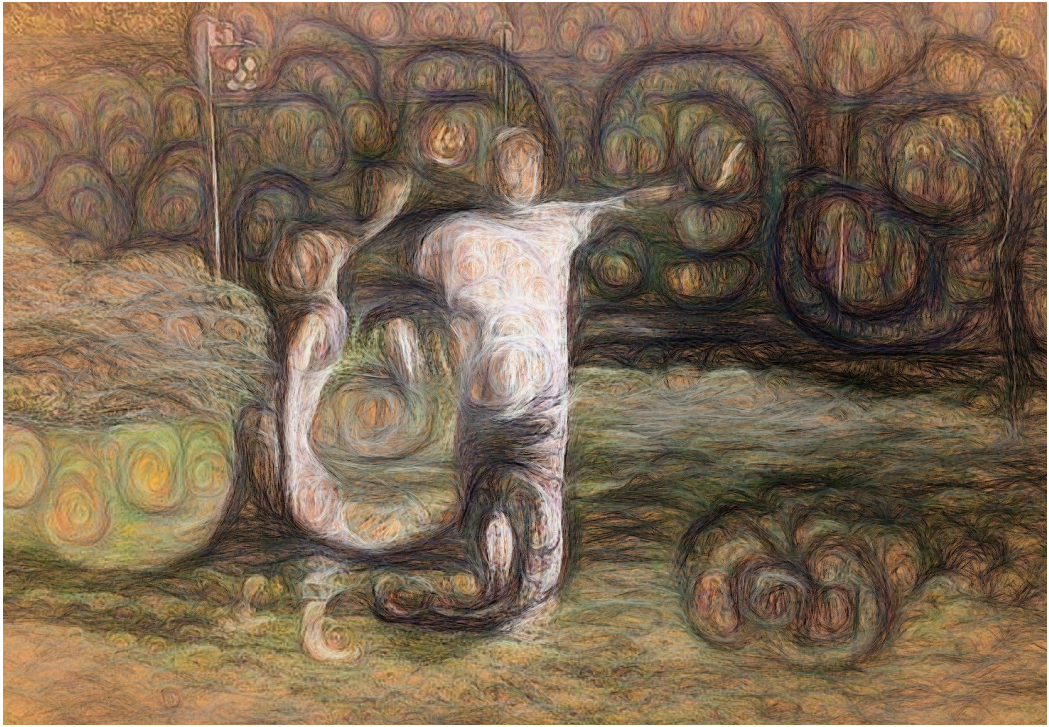

**Fig 4.** 4a: A sketch-like style was used to generate this 2D image of a 3D avatar. 4b: A still from a movie of two dancers in which Van Gogh-esque swirling style is contributed by modified DD system and hand-drawn like rendering is contributed by painterly system.

Going beyond DD's ability to *evaluate and optimize* novelty and value, we ask whether DD tends to *generate* outputs with high novelty and value. Informally, we observe that the range of outputs generated by these systems often strikes us as surprising yet visually pleasing. The viewer may be surprised not only by the form of an unexpected visual concept being "imagined" into the original image, but in the form of an unexpected manner of visual resemblance between selected inputs. Perhaps the network's support of a large number of multi-level abstracted visual features creates a rich enough space of combinations such that the optima found by DD are often not the same as a given human viewer might anticipate.

Ritchie (2007) suggests that measures more fine-grained or primitive than novelty are dissimilarity from the "inspiring set" and typicality (relative to the target art form or genre). For DD, we take the inspiring set to include both the original neural network training image set, as well as the current input image(s) to the algorithm. DD is clearly not simply replicating or near-replicating any of the training set images. By design, DD outputs bear visual similarity to the input images, but the resemblance is not so slavish as to exclude creativity.

Ritchie regards typicality as a double-edged sword: on one hand, it is an achievement for a computer to generate successful examples of a style, but on the other hand, high typicality suggests low novelty. For DD, the comparison set for typicality is not obvious (simply "contemporary art" would be one choice). Additional considerations from (Ritchie 2007) regarding repetition point to other limitations of DD's creativity. Repetition in the output arises when using DD multiple times with the same inputs and parameter settings; furthermore, there is a sense of sameness that may arise when viewing multiple DD outputs (more so when the training set and network architecture remain fixed).

Finally, DD algorithms may be particularly amenable to future extension/modifications that would enhance their ability to internally search for novelty/value. Regarding novelty, the vector spaces formed by node activations can lead to distance metrics that are more relevant and useful to human-perceived visual similarity compared to raw-pixel-based similarity. Such measures could be combined with storage of images and/or data clustering techniques to estimate the novelty of a particular generated image compared to training images or compared to a certain corpus (e.g., art of a certain genre). Regarding aesthetic value, ideas from information-based aesthetic theories (Rigau et al. 2008) such as compressibility could be applied as additional optimization constraints, using the node vectors as a basis.

### 3.2  Psychology of Creativity Concepts in AI

We now consider how this generative art program implements concepts from the psychology of creativity literature.

**Conceptual Combination and Blending** In conceptual combination or blending (Fauconnier & Turner 1998) new concepts are generated by integrating multiple pre-existing conceptual spaces. This process has been mathematically modeled (e.g., Aerts et al 2016) and has been suggested as an important mechanism underlying human creativity (Pereira & Cardoso 2002). Existing approaches to computational modeling of conceptual blending include symbolic AI-based systems (Besold & Plaza 2015) and the approach of (Thagard & Stewart 2011) which combines neural patterns using a convolution operation (different from the meaning of convolution in CNNs).

Theory and modeling of conceptual blending often includes visual blends, in which the novel output of a blending operation is expressed as an image. Visual blending is often framed in terms of combining high-level, verbalizable concepts (Martins et al. 2015; Confalonieri et al. 2015). However, the strength of DD lies in its ability to combine lower-level yet abstract image qualities such as shapes, textures and arrangements, extracting such qualities from input images without explicit labelling. Lower-level image-quality blending is likely a component of visual imagination (Richardson 2015; Heath et al. 2015) and fits the idea of preconceptual creativity (Takala 2015). We suggest that DD is a promising computational tool for exploring preconceptual visual blending and imagination which is relevant to creativity and art creation. Figure 4b shows a conceptual blend of a swirling Van Gogh style, which comes from the modified DD system, and a hand-drawn rendering, which comes from the enhanced painterly system.

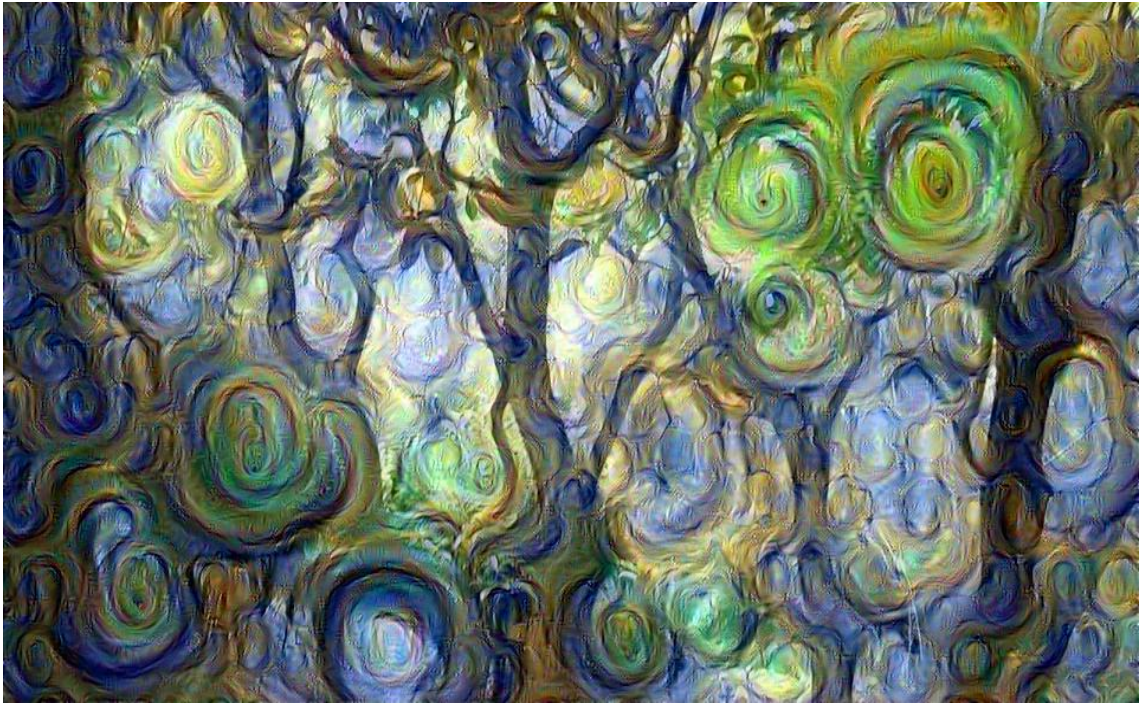

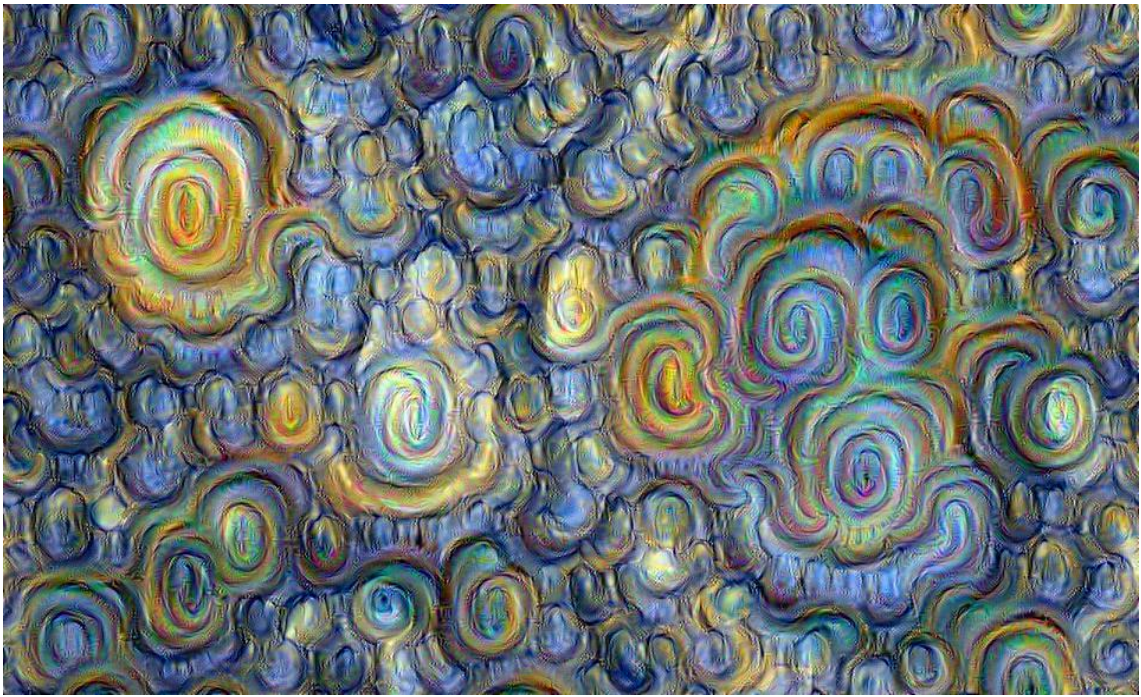

**Fig 5.** A still from a movie of two dancers in which Van Gogh-esque swirling style is contributed by modified DD system and hand-drawn rendering is contributed by painterly system.

**Contextual Focus and Dual Processing Modes**: Innovative concept combinations are thought to take place in an associative mode of thought, in which more features of the object of thought are held in mind, and thus there are more associative routes available to items that have previously gone unnoticed (Gabora 2010; Gabora and Ranjan, 2013). The ability to shift between a focused analytic mode of thought and a defocused associative mode of thought is known as contextual focus and it plays an important role in creativity (DiPaola & Gabora 2009). Elsewhere we discussed in detail how contextual focus is implemented in DD (McCaig, DiPaola, & Gabora 2016).

**Honing theory** (Gabora 2005, 2017) is a psychological theory of creativity which posits that creativity arises via cognitive restructuring interaction between an individual's worldview and the conceptual space of a task or creative problem. Honing theory (HT) predicts that creativity involves the merging and interference of memory items resulting in a single cognitive structure that is ill-defined and can be said to exist in a state of potentiality, and which can be formally described as a superposition state. The idea becomes increasingly well-defined and transforms from potential to actual through interaction with internally and externally generated contexts. The idea could actualize in different ways depending on the contexts the idea interacts with, or perspectives it is viewed from. Several aspects of HT are mirrored in the mechanisms of the DeepDream algorithm, making it an interesting line of inquiry to compare the two frameworks. We began this examination in our previous paper (McCaig, DiPaola, & Gabora 2016). Here we elaborate on parallels between HT and DD in two categories/themes:

**1. Iteratively exploring a space of potential associations:** In HT, latent associations exist as overlapping neural cell assemblies that may be triggered due to different ways of viewing a task through the lens of past experiences. In DD, we can view the neural net feature activations as a direct model of HT's distributed neural cell assemblies. In DD's guided-dream mode of operation, we can view the guide image(s) as part of HT's "memory of previous experiences" while the source image plays the role of the current problem demanding a creative solution. The neural representations stemming from different areas of texture and form within the guide image are explicitly tested for overlap/resonance with neural representations of the source image. This mirrors the HT process in which different associations can be made between a problem at hand and previously formed perspectives/memories which have some connection or overlap to the problem. Aspects of the source image that bear stronger resonance with certain guide qualities are iteratively enhanced during the dream process, moving from mere potential associations to strong new themes and image qualities in the eventual created output image.

There is a second way to map HT onto DD operation, regardless of whether a guide image is used: viewing the source image's different encodings in higher (more abstract) or lower (less abstract) network layers as alternate perspectives (neural cliques) that can potentially push the creation of an output image towards different visual stylistic directions. The notion that neural cliques vary in their degree of abstractness/generality is discussed within HT (Gabora 2010). The role of different neural levels of abstraction in visual art creativity remains a rich topic for further exploration within both DD and HT frameworks.

**2. Honing as Problem/Worldview interaction:** Central to HT is the view that people work on a creative problem by modifying both the problem representation and their worldview itself. If we draw a parallel between worldview and the structure of the network itself in DD, we see that training the network on new task-related data, such as our Art Database, is a type of honing process. In this process the neural-code representations of both the new training material and previously understood image material are dynamically changed.

In the typical usage of DD, such "honing-during-training" takes place as a precursor to the generation of actual particular artwork outputs. However, we can also point to a more dynamic real-time honing process that takes place within the iteration/association loop of a certain image production. An exciting element of working with DD as an artist is seeing the honing process unfold as multiple test-images are progressively produced. We have created an interactive artwork which puts DD in dialog with HT by highlighting both "honing-during-training" and also giving the audience dynamic control over the "honing-during-generation" process (see Fig. 5).

Using our new hierarchical stochastic tiling method, we trained work with a subset of more modern (abstraction) art and guided it increasingly toward a Van Gogh "Starry Night" style, and this computational honing process was used to make the interactive piece. Figure 5 shows captured stills from this work captured at two different abstraction levels. Using intuitive body-motion controls the audience can dynamically steer the live animated art piece through different levels of stylized abstraction, experiencing the honing of art towards a certain style in real-time. Although it is difficult to convey this shifting between potentiality and actualization with static images, it is very evident in as one moves through this interactive artwork depicted in these two stills.

Notably, in honing-during-generation with DD, the network weights themselves are typically not altered. According to the HT notion of art-creation as struggling with or even "healing" the artist's worldview, we might obtain more cognitively valid, or artistically interesting results by incorporating "training inside the loop," where the neural network structure and/or weights would be modified even within the generation process of each individual image artwork. This is an appealing direction for future experimentation.

**Personal Style:** Personal style is a key defining aspect of a creator's sense of identity. Figure 6a is an example of how the same source can be reimagined, here in multiple different interpretations, i.e. clockwise from upper left, us nature, fire, water and universe. Indeed, it has been shown empirically that we each have a personal creative style that is recognizable by others, and this recognizability even extends across domains (e.g., creative writing students who are familiar with one anothers' writing style can recognize at significantly above chance levels which work of art was done by which student (Gabora, O'Connor, & Ranjan, 2012; Ranjan, 2014).

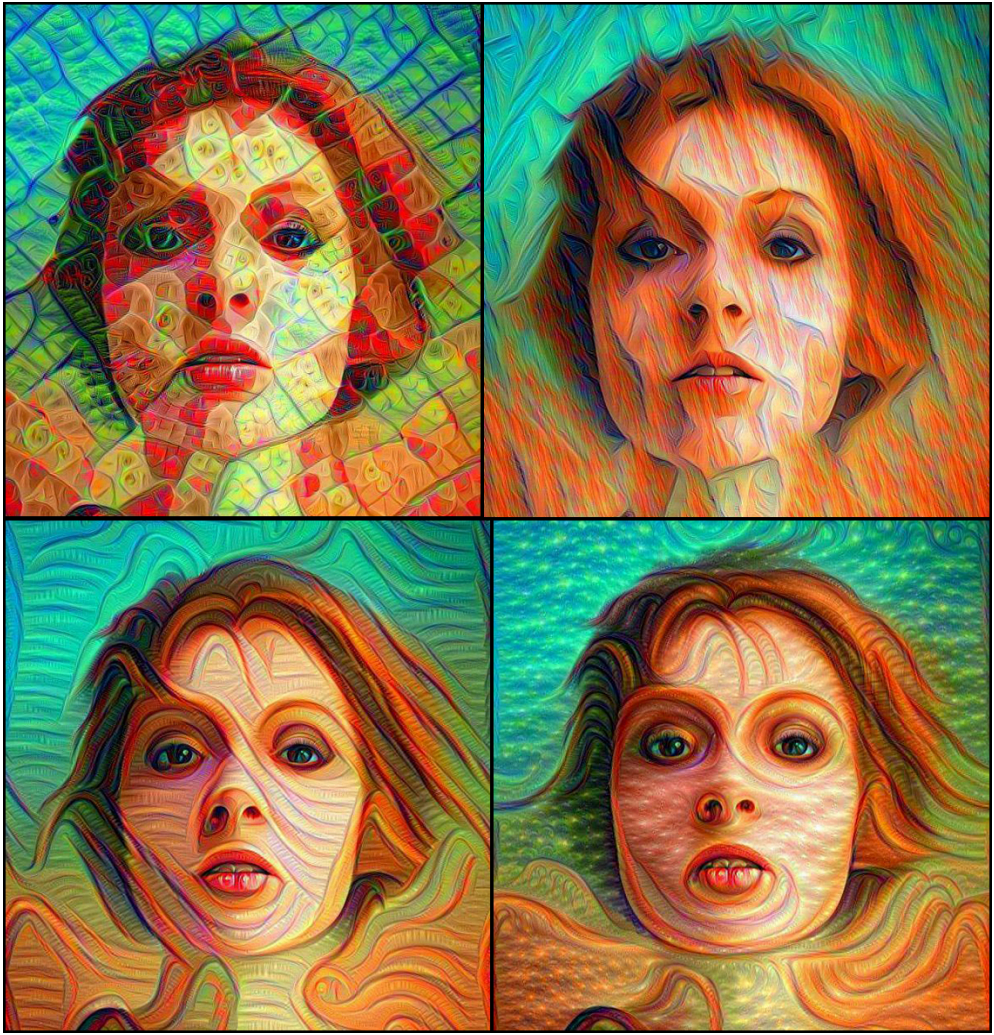

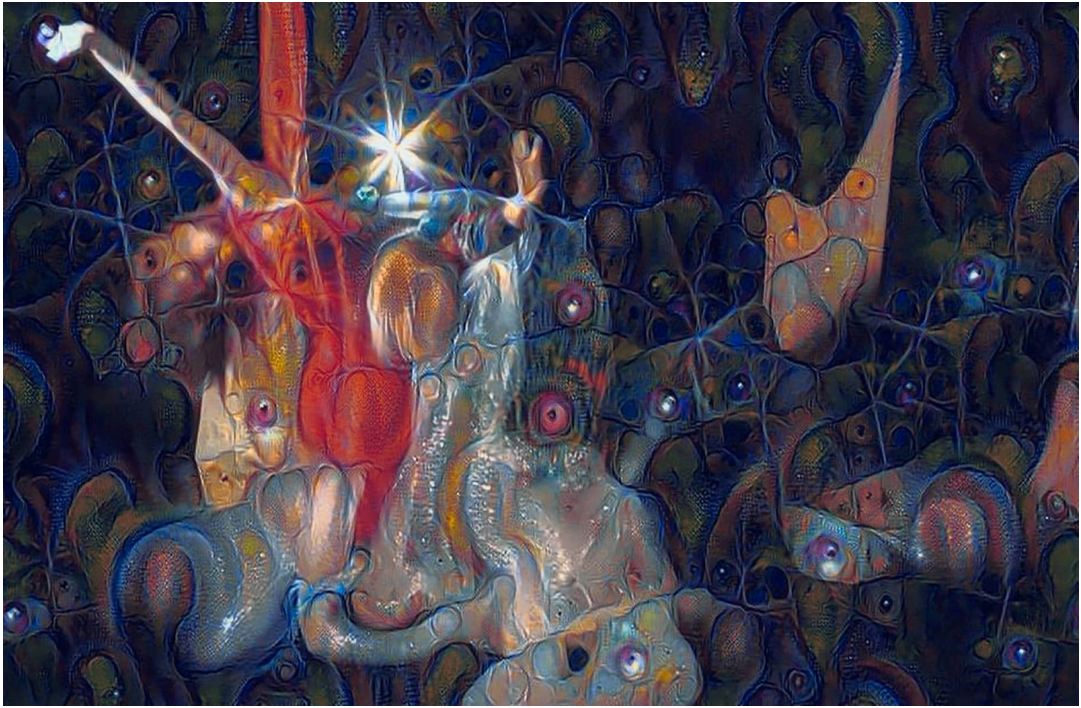

**Fig 6.** 6a: The same image rendered in four different styles. 6b: Combining Van Gogh's style with the organic style to create a "biological spore" look in this dancer movie still.

The Stochastic Tiling Technique makes it possible to merge different styles to generate a unique emergent new style. The still from a movie of two dancers, in which the swirling Van Gogh style was combined with the hand-drawn rendering style (Figure 3), demonstrates how a new style can emerge through the personalized tuning process that incorporates elements of previously learnt styles. As another example of this, in Figure 6b, the Van Gogh style is merged with the organic style creating a "wet biological spore" look.

### 4    Concepts from the Psychology of Creativity that Currently Elude Algorithmization

We now look at concepts that come out of the psychological literature on creativity that (as of yet) have not found a place in computational models of creativity.

**Intrinsic Motivation:** While human creativity is typically thought as extrinsically motivated – that is the act of creation illicits an object, reward or achievement of some valued outcome, it is also sometimes intrinsically motivated, where humans create because we derive pleasure from the act of creating (Amabile, 1983). It remains an open question whether or not the extrinsic versus intrinsic motivation issue can be addressed using computational creativity.

**The "Gap":** Creativity is often thought to originate with what is sometimes referred to as a gap, or sense of incompletion. It may arise spontaneously, or slowly over the course of years, and be trivial (e.g., drawing a squiggle and wondering what additional doodling would make the squiggle appealing) or of worldly consequence. Although it seems more intuitively straightforward to see how computational creativity programs could incorporate a gap, the concept is not widely used in the computational creativity literature.

**Therapeutic Impact of Creativity:** Creative activity can have a therapeutic impact on the creator (Forgeard, 2013). It can be intrinsically rewarding (Kounios & Beeman, 2014;). Although the creative process may at times be frustrating and draining, or involve working through negative material, there is evidence that high levels of creativity are correlated with positive affect (Hennessey & Amabile, 2010), and the ability to manage intense feelings (Moon, 1999). Clinical practitioners of art therapy note that imagery and creative engagement can deepen communication between client and therapist (Moon, 2009). Art therapy can also enhance self-understanding and facilitate the process of finding healthier ways of handling situations and interacting with others (Riley, 1999). It can provide access to issues that are difficult to verbalize, and either bring them to the surface in a nonverbal form, or provide a springboard for discussion (Malchiodi, 2007). There is also evidence that creativity can enhance ones' sense of self (Garailordobil & Berrueco, 2007;). Is there any sense in which such therapeutic effects of creative immersion are achievable by a computer? This is not a trivial question, as for many creative people such therapeutic benefits are the whole reason they engage in creative activities in the first place. They are the raison d'être behind the massively popular creative therapies—also called expressive therapies—such as art therapy, music therapy, dance therapy, and writing therapy.

Even if it were possible in theory to simulate therapeutic benefits of creativity in a computer, it may be that some of these therapeutic benefits cannot be achieved unless the creative program is embedded in an artificial society of similar agents. The therapeutic effects of creativity may stem from its capacity to enhance feelings of connection to, and appreciation by, others. Clearly computational creativity is a long way from this interpersonal therapeutic benefit, but it can nevertheless serve as a beacon to inspire further work.

## 5 Conclusion and Future Directions

We have outlined a current effort in generative art focusing on the DeepDream algorithm as an example of how computational creativity is benefiting from cross-fertilization with the cognition of creativity literature, and how it may continue to do so. We suggest that deep neural networks (with accompanying search/optimization algorithms) to produce creative visual blends and artifacts have entered a new and promising phase, both for models of creativity and for practical art-generating systems. We have shown that in a number of respects these advances align nicely with theoretical work from the psychology of creativity, while in other respects ideas about how the creative process works loom as factors to consider and potentially inspire new developments.

**Acknowledgments**

This research was supported in part by a grant to Gabora from the Natural Sciences and Engineering Research Council of Canada and SSHRC grant to DiPaola.